\documentclass[10pt,twocolumn,letterpaper]{article}

\usepackage{cvpr}              

\usepackage{listings}
\usepackage{algorithmicx}
\usepackage{color}
\usepackage{soul}
\usepackage{adjustbox}
\usepackage{bm}
\usepackage{algorithmicx}
\usepackage{algorithm}
\usepackage{algpseudocode}
\usepackage{pifont}
\usepackage{wasysym}
\usepackage{placeins} 
\usepackage{graphicx}
\usepackage{multirow}
\usepackage[hang,flushmargin]{footmisc}
\usepackage{balance}
\usepackage{latexsym}
\usepackage[most]{tcolorbox}
\usepackage{float}

\lstdefinestyle{cuda}{
	belowcaptionskip=1\baselineskip,
	breaklines=true,
	xleftmargin=\parindent,
	language=C,  
	showstringspaces=false,
	basicstyle=\footnotesize\ttfamily,
	keywordstyle=\bfseries\color{PineGreen},
	commentstyle=\color{white!40!black},
	identifierstyle=\color{NavyBlue},
	stringstyle=\color{orange},
	escapeinside={(*@}{@*)},
	morekeywords={shared, is, to},
	numbers=left,
	morekeywords={[2]{copy,syncthreads,nnz}},
	keywordstyle={[2]{\color{Purple}}},
}     

\usepackage{tikz}
\newcommand*\Circled[2][gray!40]{
	\tikz[baseline=(char.base)]{\node[
        shape=circle, draw=none,  thick, 
        fill=#1 ,inner sep=0.9pt] (char) 
    {\textcolor{black}{#2}}; 
}}

\usepackage{framed}

\definecolor{that-yellow}{HTML}{ddc26f}

\definecolor{first}{HTML}{FF9999}
\definecolor{second}{HTML}{FFCC99}
\definecolor{yellow}{HTML}{FFF8AD}
\definecolor{green}{HTML}{85D0FF}
\newtcbox{\topone}{on line, colback=first, colframe=first, boxrule=0pt, arc=0pt, boxsep=1pt, left=2pt, right=2pt, top=1pt, bottom=1pt}
\newtcbox{\toptwo}{on line, colback=second, colframe=second, boxrule=0pt, arc=0pt, boxsep=1pt, left=2pt, right=2pt, top=1pt, bottom=1pt}
\newtcbox{\greenbar}{on line, colback=green, colframe=green, boxrule=0pt, arc=0pt, boxsep=1pt, left=2pt, right=2pt, top=1pt, bottom=1pt}

\newcommand{\barChartA}[1]{%
  \smash{\pgfmathsetmacro{\barlen}{#1/4*0.8}
  \begin{tikzpicture}[baseline=(base)] 
    \node[inner sep=0pt] (base) at (0.9cm, 0.053cm) {};
    \fill[green] (0,0) rectangle (\barlen cm, 0.37cm);
    \node[anchor=center] at (0.45cm, 0.18cm) {#1};
  \end{tikzpicture}%
}}

\newcommand{\barChartB}[1]{%
  \pgfmathsetmacro{\barlen}{#1/4*0.7}
  \smash{%
    \begin{tikzpicture}[baseline=(base)] 
      \node[inner sep=0pt] (base) at (0.9cm, 0.053cm) {};
      \fill[green] (0,0) rectangle (\barlen cm, 0.37cm);
      \node[anchor=center] at (0.45cm, 0.18cm) {\textbf{#1}};
    \end{tikzpicture}%
  }%
}

\makeatletter
\let\@authorsaddresses\@empty
\makeatother

\definecolor{cvprblue}{rgb}{0.21,0.49,0.74}
\usepackage[pagebackref,breaklinks,colorlinks,allcolors=cvprblue]{hyperref}

\def\paperID{****} 
\def\confName{CVPR}
\def\confYear{2026}

\title{Gaussians on a Diet: High-Quality Memory-Bounded 3D Gaussian Splatting Training}

\author{Yangming Zhang, Jian Xu\\
Dept. of Computer Science\\
University of Texas at Arlington\\
{\tt\small \{yxz0925, jxx3451\}@mavs.uta.edu}
\and
Chaojian Li\\
School of Computer Science\\
Georgia Institute of Technology\\
{\tt\small cli851@gatech.edu}
\and
Kunxiong Zhu, Wei Niu, Gagan Agrawal\\
School of Computing\\
University of Georgia\\
{\tt\small \{kz96891, wniu, gagrawal\}@uga.edu}
\and
Yang Katie Zhao\\
Department of ECE\\
University of Minnesota - Twin Cities\\
{\tt\small yangzhao@umn.edu}
\and
Jian Wang\vspace{2.5mm}\\
Snap Inc.\vspace{2.5mm}\\
{\tt\small jwang4@snap.com}
\and
Yingyan Celine Lin\\
School of Computer Science\\
Georgia Institute of Technology\\
{\tt\small celine.lin@gatech.edu}
\and
Miao Yin$^\dagger$\\
Dept. of Computer Science\\
University of Texas at Arlington\\
{\tt\small miao.yin@uta.edu}
}

\begin{document}
\twocolumn[{
\maketitle
  \centering
   \vspace{-5mm}
  \begin{minipage}[t]{0.59\textwidth}
    \centering
    \includegraphics[height=5cm]{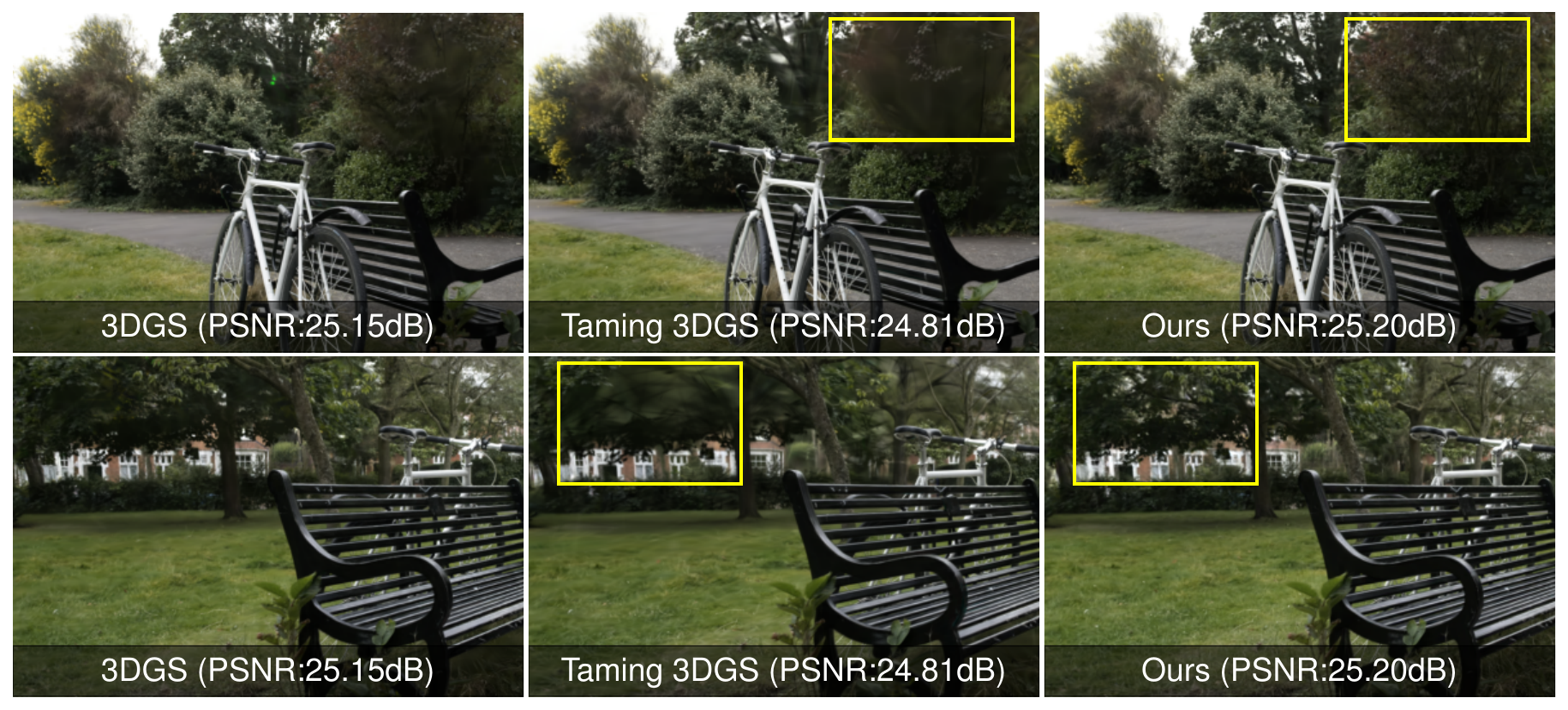}
    \label{fig:teaser_rendered_image1}
  \end{minipage}
  \hspace{0\textwidth} 
  \begin{minipage}[t]{0.4\textwidth}
    \centering
    \includegraphics[height=5cm]{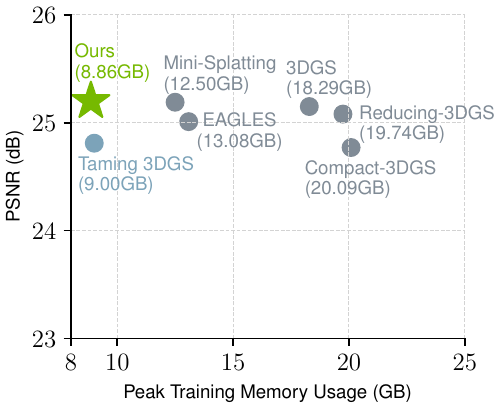}
    \label{fig:teaser_rendered_image2}
  \end{minipage}
  \vspace{-7.2mm}
  \captionof{figure}{We present a memory-bounded 3D Gaussian Splatting training framework, enabling lower peak training memory and higher rendering quality, compared to existing state-of-the-art methods.}
  \label{fig:teaser_rendered_image_both}
  \vspace{2.5mm}
}]

\begin{abstract}
\let\thefootnote\relax\footnotetext{$^\dagger$Corresponding author.}3D Gaussian Splatting (3DGS) has revolutionized novel view synthesis with high-quality rendering through continuous aggregations of millions of 3D Gaussian primitives. However, it suffers from a substantial memory footprint, particularly during training due to uncontrolled densification, posing a critical bottleneck for deployment on memory-constrained edge devices. While existing methods prune redundant Gaussians post-training, they fail to address the peak memory spikes caused by the abrupt growth of Gaussians early in the training process. To solve the training memory consumption problem, we propose a systematic memory-bounded training framework that dynamically optimizes Gaussians through iterative growth and pruning. In other words, the proposed framework alternates between incremental pruning of low-impact Gaussians and strategic growing of new primitives with an adaptive Gaussian compensation, maintaining a near-constant low memory usage while progressively refining rendering fidelity. We comprehensively evaluate the proposed training framework on various real-world datasets under strict memory constraints, showing significant improvements over existing state-of-the-art methods. Particularly, our proposed method practically enables memory-efficient 3DGS training on NVIDIA Jetson AGX Xavier, achieving similar visual quality with up to 80\% lower peak training memory consumption than the original 3DGS. 
\end{abstract}    
\section{Introduction}

3D Gaussian Splatting (3DGS) \cite{kerbl20233d} has recently emerged as a powerful paradigm for novel view synthesis and 3D reconstruction \cite{chen2024survey}. By representing a scene as a set of 3D Gaussians, each with parameters such as spatial position, scale, opacity, rotation, and spherical harmonic (SH) coefficients for view-dependent color. 3DGS enables differentiable rendering with promising visual quality. This approach has demonstrated state-of-the-art performance in rendering speed and quality, achieving immersive view synthesis at high resolutions in real time. However, these gains come at a substantial memory cost -- 3DGS models often employ millions of Gaussians for a single scene, leading to significant memory consumption \cite{bagdasarian20243dgs}. This reliance on a large number of primitives not only inflates the model size but also restricts deployment on edge devices or other memory-constrained platforms \cite{morgenstern2024compact,yu2024cogs}. In practice, the heavy memory footprint of 3DGS-based models has become a key bottleneck, limiting their scalability and adoption in resource-limited settings.

\begin{figure}[t]
    \centering
    \includegraphics[width=1\linewidth]{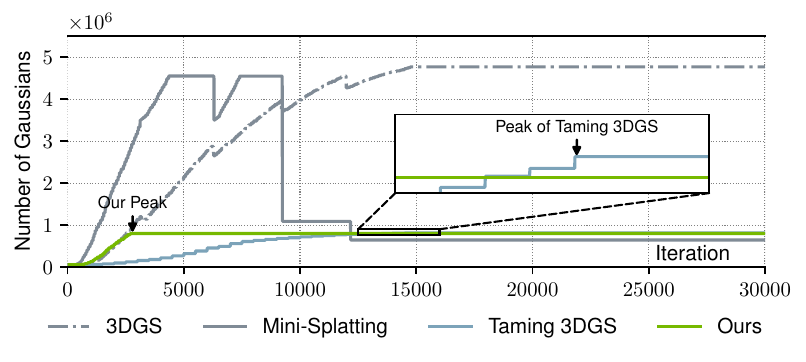}
    \vspace{-9mm}
    \caption{\textbf{The variation curves for the numbers of Gaussians over iteration.} Our method iteratively grows and prunes Gaussians under the memory constraint, while 3DGS \cite{kerbl20233d} and Mini-Splatting \cite{fang2024mini} densify Gaussians to millions and remove them afterwards. }
    \label{fig:numbers_comparison}
    \vspace{-2mm}
\end{figure}

Existing works \cite{deng2024efficient,fang2024mini,niemeyer2024radsplat,papantonakis2024reducing,zhang2024lp} mainly focus on pruning redundant Gaussians to obtain a compact scene representation. Nevertheless, the challenging problem is that existing pruning approaches are applied after the uncontrollable densification process in the original 3DGS training framework \cite{fan2023lightgaussian,rota2024revising,kheradmand20243d,yu2024mip, lee2024deblurring}, where the Gaussian primitives suddenly expand to a tremendous number as shown in Fig. \ref{fig:numbers_comparison}, e.g., several million for the bycicle scene, leading to a substantial peak training memory consumption (refer to Fig. \ref{fig:teaser_rendered_image_both}\footnote{Here, we report our peak memory usage on GTX 4090 without engineering optimization for a fair comparison, while Mini-Splatting \cite{fang2024mini} leverages extra compression by downgrading the orders of SH coefficients to one before pruning to reduce memory. }). Even though those methods successfully reduce the memory footprints in the rendering phase, the peak memory consumption is significantly higher than the memory size of edge systems, thereby hindering real-time 3D applications in real-world settings \cite{ matsuki2024gaussian}.

Despite the practical significance of peak training memory usage in 3DGS, this issue remains understudied. Prior work \cite{mallick2024taming} mitigates memory spikes by regulating Gaussian growth via a predictable curve and selectively cloning/splitting primitives using a computationally intensive importance score. While this approach reduces peak Gaussian counts, it suffers from some key drawbacks. (1) Strict growth restrictions in early training stages limit representational capacity, leading to accumulated rendering errors that persist due to insufficient dynamic refinement. (2) Cloned Gaussians inherit identical initial positions, resulting in redundant gradient updates that fail to effectively capture missing scene details. (3) The absence of an active Gaussian removal mechanism allows poorly optimized (``ill'') Gaussians to persist throughout training.

To address those limitations and practically enable real-time 3DGS training on memory-constrained devices, in this paper, we conduct an in-depth study on the unsatisfactory performance of existing training approaches.  Our motivation is inspired by the Lottery Ticket Hypothesis \cite{frankle2018lottery},which suggests the existence of a sparse subnetwork capable of achieving performance comparable to a dense one. We extend this hypothesis to 3DGS, whether an optimal sparse Gaussian model can be trained from scratch under strict memory limitations. Building on this insight, we propose a systematic memory-bounded 3DGS training framework based on dynamic growing and removal of Gaussian primitives, which can strictly satisfy the practical memory constraints. Our proposed training framework alternatively identifies, grows, and prunes  Gaussian primitives in every predefined iteration (e.g., 100), where ``ill'' Gaussians are dynamically deleted and ``healthy'' Gaussians are subsequently regenerated. This iterative process ensures consistently low memory usage while discovering effective primitives that match the rendering quality of the original 3DGS at significantly reduced training memory consumption. As illustrated in Fig. \ref{fig:teaser_rendered_image_both}, our dynamic strategy outperforms one-shot pruning \cite{fang2024mini,girish2024eagles} in both memory efficiency and visual fidelity.

\section{Related works}

\subsection{Novel View Synthesis}
Novel View Synthesis (NVS) \cite{mildenhall2021nerf} aims to generate photorealistic images of a scene from previously unseen viewpoints, given a set of input images captured from known camera poses. As a fundamental problem in computer vision and graphics, NVS is critical to a wide range of applications, including virtual reality, robotics, digital content creation, and autonomous navigation.

Prior advances in neural rendering, particularly Neural Radiance Fields (NeRF) and its successors \cite{barron2022mip,barron2023zip,navaneet2024compgs,roessle2022dense,zhang2024fregs}, have significantly improved the fidelity of novel view synthesis. These methods represent scenes as continuous volumetric functions and are trained to reproduce the appearance of input images from arbitrary viewpoints. While NeRF-based techniques achieve impressive visual quality, they typically require dense sampling and long training times, making them less practical for real-time or resource-constrained scenarios \cite{barron2021mip}. Recently, as an explicit point-based rendering approach, 3D Gaussian splatting \cite{kerbl20233d,lu2024scaffold}, which aims to strike a better balance between rendering quality, speed, and memory efficiency, has emerged as an alternative for NeRF and a promising direction for real-time novel view synthesis, leveraging compact and differentiable point representations to achieve high visual fidelity with fast rendering performance \cite{fei20243d}.

\subsection{Compact 3D Gaussian Splatting}

Although 3DGS achieves significant progress in photorealistic scene representation and novel view synthesis, its reliance on millions of primitives creates significant memory bottlenecks that hinder practical deployment   \cite{bagdasarian20243dgs,hanson2024speedy,lee2025optimized}. This challenge has led to growing interest in compact 3DGS methods, which aim to preserve rendering fidelity while significantly reducing the number of primitives \cite{bao20253d,liu2024maskgaussian,ye20243d}. For instance, LightGaussian \cite{fan2023lightgaussian} reduces final storage by pruning redundant Gaussians based on a global importance score after training, while RadSplat \cite{niemeyer2024radsplat} improves pruning robustness by replacing the sum with a max operator for score computation.

Although these post-training pruning strategies reduce memory usage during inference, they do not alleviate the high peak memory consumption incurred during training \cite{feng2024flashgs,navaneet2024compgs,zhang2024gaussian}. To address this, Taming 3DGS \cite{mallick2024taming} introduces a steerable densification mechanism that selectively densifies impactful Gaussians, enabling a more predictable and memory-aware growth trajectory. However, its infrequent densification and slow growing speed result in suboptimal and poorly positioned Gaussians remaining in the model, limiting the representation details and overall rendering quality. Based on the in-depth study on the limitations of \cite{mallick2024taming}, our work progressively refines the model via iteratively growing and pruning, dynamically preserving most ``healthy'' Gaussians under memory bounds.

\section{Background and Motivation}

\subsection{Background of 3D Gaussian Splatting}

3D Gaussian Splatting (3DGS) \cite{kerbl20233d} represents the scenes using an optimized collection of anisotropic 3D Gaussians. Each Gaussian $G$ is defined by its covariance matrix $\bm{\Sigma}$ and center position $\bm{\mu}$ as
\begin{equation}\label{eqn:gaussian-defination}
  G(\bm{x})=\exp\left(-\frac{1}{2}(\bm{x}-\bm{\mu})^\top\bm{\Sigma}^{-1}(\bm{x}-\bm{\mu})\right),
\end{equation}
where $\bm{x}$ is an arbitrary position in the 3D scene. The covariance matrix $\bm{\Sigma}$ is generally decomposed as $\bm{\Sigma}=\bm{R}\bm{S}\bm{S}^\top\bm{R}^\top$, where $\bm{R}$ is a rotation matrix and $\bm{S}$ is a diagonal scaling matrix. 

To render a 2D image from the 3D scene, 3DGS projects 3D Gaussians onto the image plane based on the camera parameters. The projected 2D covariance matrix is computed as $\bm{\Sigma'}=\bm{J}\bm{W}\bm{\Sigma}\bm{W}^\top\bm{J}^\top$
where $\bm{W}$ represents the view transformation matrix, and $\bm{J}$ is the Jacobian of the affine approximation of the projective transformation. Then, the final color $C$ at each image pixel is computed by blending all $N$ depth-ordered Gaussians contributing to the pixel as
\begin{equation}\label{eqn:render-color}
  C=\sum_{i\in N}c_i\alpha_i \prod_{j=1}^{i-1}(1-\alpha_j).
\end{equation}
Here, $c_i$ is the color of each Gaussian derived from the SH coefficients. $\alpha_i$ is the ray transmittance calculated by overlapped Gaussians' opacity $o_i$ and the relative distance between rendered pixel position $x$ and 2D view-plane Gaussian's center $\mu_i$ \cite{niemeyer2024radsplat}, i.e.,
{\footnotesize
\begin{equation}\label{eqn:alpha}
\alpha_i(\bm{x})=o_i \exp \left(-\frac{1}{2}\left(\bm{x}-\mathcal{R}\left(\boldsymbol{\mu}_i ; \theta\right)\right)^{\top} \mathcal{R}_{\theta}\left(\boldsymbol{\Sigma}_i\right)^{-1}\left(\bm{x}-\mathcal{R}\left(\boldsymbol{\mu}_i ; \theta\right)\right)\right)
\end{equation}
}
where $\theta$ is the camera pose and $\mathcal{R}$ is the 3D-to-image plane projection operation.

During training, the Gaussians are initialized from a sparse point cloud generated by Structure-from-Motion (SfM) \cite{schonberger2016structure}. Then, each Gaussian attribute is optimized with the gradient backpropagation to minimize the reconstruction error.

\begin{figure}[t]
    \centering
    \includegraphics[width=\linewidth, height=5.2cm, keepaspectratio]{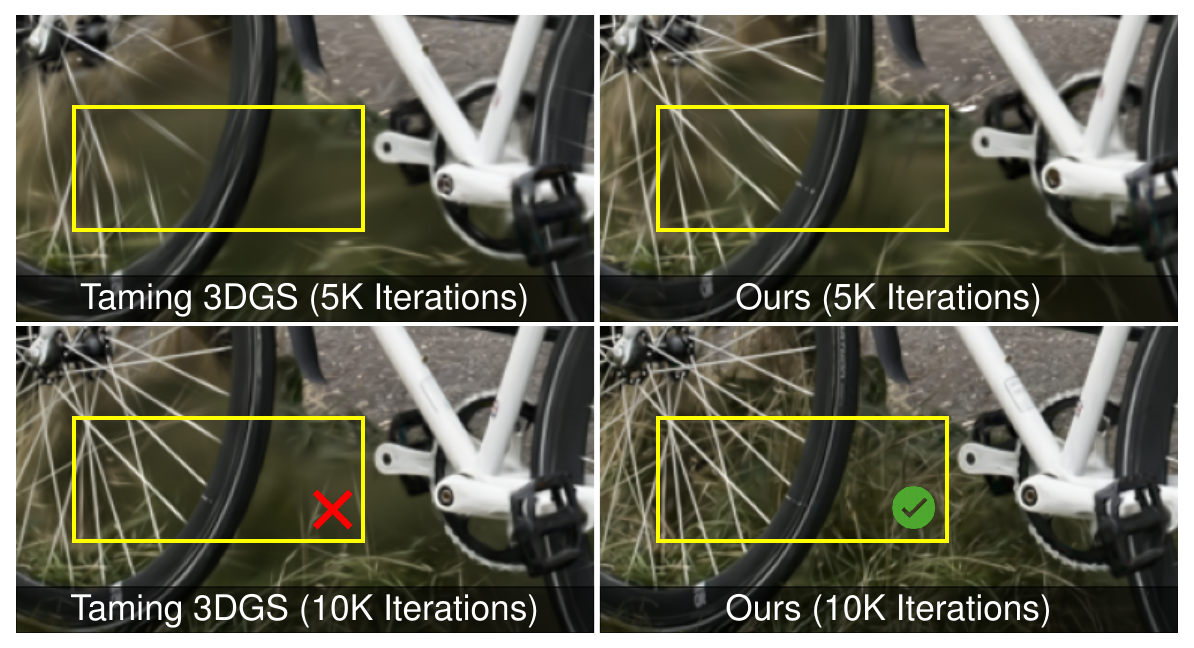}
     \vspace{-8mm}
    \caption{The ``error'' information (grass behind the bicycle) cannot be corrected after long-term iterations by Taming 3DGS \cite{mallick2024taming}, while our proposed solution can effectively optimize those areas.}
    \label{fig:convergence_speed}
    \vspace{-2mm}
\end{figure}

\subsection{Motivation for Resolving Problems in 3DGS Training}
\label{sec:motivation}
We conduct an in-depth study on existing training approaches, including the original 3DGS \cite{kerbl20233d} and subsequent pioneering works \cite{mallick2024taming}, which propose the accelerated 3DGS training. Our analysis highlights significant limitations in these methods, motivating the development of a new training framework that dynamically grows and removes Gaussians based on more effective criteria. Below, we present three main explorations of existing approaches.

Although these post-training pruning strategies reduce memory usage during inference, they do not alleviate the high peak memory consumption incurred during training \cite{feng2024flashgs,navaneet2024compgs,zhang2024gaussian}. To address this, Taming 3DGS \cite{mallick2024taming} introduces a steerable densification mechanism that selectively densifies impactful Gaussians, enabling a more predictable and memory-aware growth trajectory. However, its infrequent densification and slow growing speed result in suboptimal and poorly positioned Gaussians remaining in the model, limiting the representation details and overall rendering quality (see Fig. \ref{fig:convergence_speed}). Based on the in-depth study on the limitations of \cite{mallick2024taming}, our work progressively refines the model via iteratively growing and pruning, dynamically preserving most ``healthy'' Gaussians under memory bounds.
\begin{figure}[t]
    \centering
    \includegraphics[width=1\linewidth]{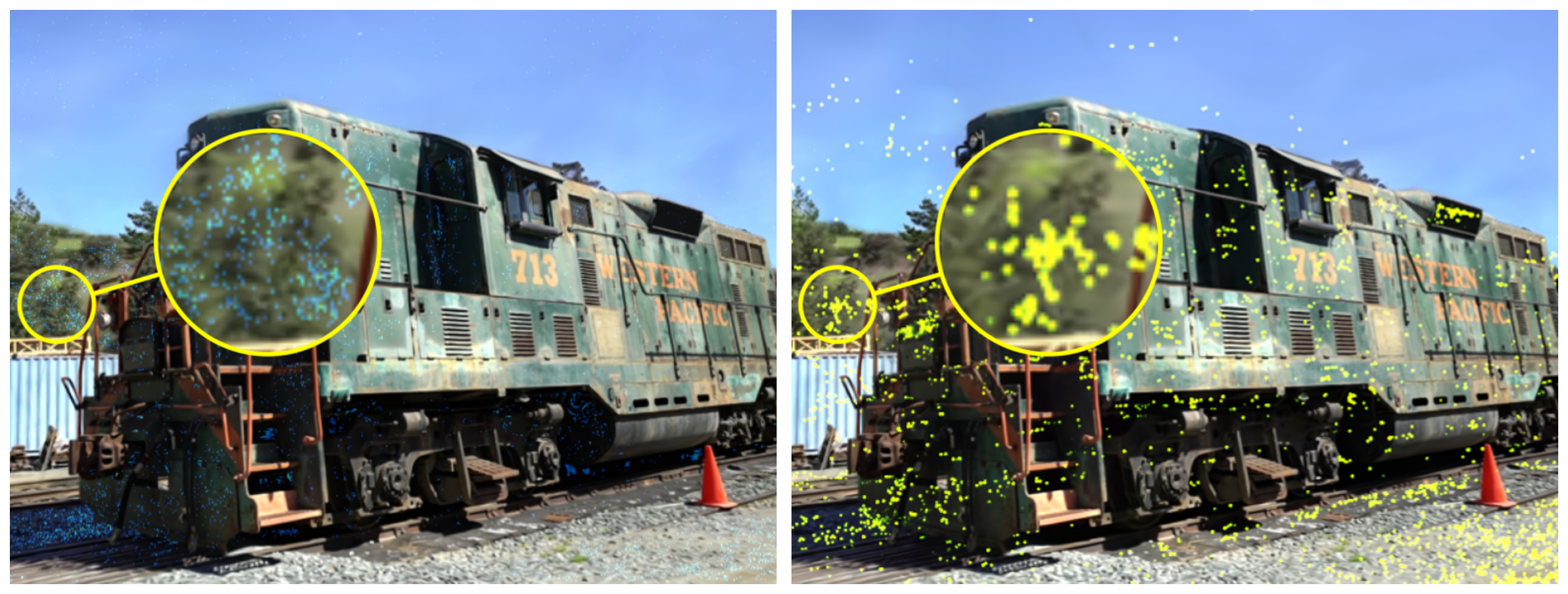}
    \vspace{-8mm}
    \caption{\textbf{Illustration of Gaussian compensation.} (left) Color gradient per pixel. (Right) Compensated Gaussians in yellow color. Our compensation step recovers the high-frequency region that is hard to capture by the original densification (e.g., rubble under the train).}
    \vspace{-3mm}
    \label{fig:compensate_gaussians}
\end{figure}

\noindent\textit{\Circled{1} \ul{Existing strategies cannot adjust Gaussian primitives dynamically, leading to accumulated error.}} 
The densification process in 3DGS \cite{kerbl20233d} is governed by the adaptive density control, which operates on a predetermined schedule. During the densification, the positional gradient magnitude for each Gaussian is tracked and averaged over all rendered views, resulting in a score. If it exceeds a user-defined threshold, the primitive is considered for growth through either cloning or splitting, depending on its size as determined by the scaling matrix.
To solve the uncontrollable number of Gaussians and the challenges in threshold determination, the subsequent methods \cite{mallick2024taming,rota2024revising} design a parabolic curve to define a schedule of new primitives at each step. Based on the predictable densification curves, they add new Gaussians by cloning or splitting existing ones according to the developed complex importance and error-correction scores. 

Even though the above methods can effectively regulate the number of Gaussians, they grow the Gaussians slowly and achieve the user-specified budget after a long-term period, i.e., 15,000 iterations. This growing strategy limits the representation power due to the limited number of Gaussians before reaching the budget, leading to a performance drop. Additionally, existing strategies prune unimportant Gaussians based on opacity \cite{lu2024scaffold} or other criteria \cite{lee2024compact} in every long-term iteration, e.g., 500 iterations, which fails to figure out the truly important Gaussians. This lies in that after 500 iterations, the ``error'' information is mixed among all Gaussians, making it challenging to remove the redundancy without frequent operation on those Gaussians. As shown in Fig. \ref{fig:convergence_speed}, with the two main limitations, existing methods cannot correctly optimize Gaussians under a bounded number of Gaussians.

\textbf{Proposed solution:} Inspired by the sparse training method \cite{frankle2018lottery,han2016dsd} for under-parameterized neural networks, we propose a dynamic approach that grows and removes Gaussians frequently (e.g., every 50 iterations) and adaptively. In our proposed training framework, at every step, we recognize a small proportion of ``ill'' Gaussians based on the blending weight (which will be introduced in the next Section) instead of opacity. Then, after removing those ``ill'' Gaussians, we add the same size of new Gaussians in the needed area. With this adaptive strategy, we can keep a sufficient capacity of Gaussians at the beginning without crafted growing curves. 
Our results show that our dynamic method can find the ``healthy'' Gaussians and remove the redundancy effectively, outperforming existing work \cite{mallick2024taming} with fewer iterations.

\noindent\textit{\Circled{2} \ul{Naive clone or split limits the representation.}} Following the 3DGS densification, prior methods  \cite{kerbl20233d,zhou2024feature} clone or split Gaussian primitives according to positional gradients, and the added Gaussians overlap the original ones with similar parameters. This naive operation causes cloned Gaussians to receive similar gradient updates during optimization. The similarity in gradients hinders their ability to diverge spatially, leading them to remain overlapped for extended periods \cite{deng2024efficient}. Consequently, a significant number of low-opacity Gaussians persist in the scene, leading to redundancy that is challenging to mitigate while increasing computational and memory overhead. Besides, due to the lack of pixel-wise information, existing methods cannot deal with areas where more Gaussians are needed but gradients fail to recognize, leading to permanently low-quality rendering results.

\textbf{Proposed solution:} To resolve this issue, the densification process needs additional information to provide randomness, but the introduced information should have a negligible impact on the rendering results. In our proposed framework, we shift the added Gaussians by a small distance based on the accumulated positional gradient, reducing the overlap between the two primitives. In this way, those Gaussians can receive gradually distinct gradient updates, thereby the new Gaussians will move to the correct location adaptively, which benefits the rendering quality. In other words, our proposed solution strengthens the robustness of the scene representation. Additionally, we propose a pixel-wise Gaussian compensation method performed at the end of each step. Specifically, we select pixels with the highest losses and put extra Gaussians on the rays rendering those pixels. As shown in Fig. \ref{fig:compensate_gaussians}, our results show that our compensation identifies and recovers the high-frequency region that is hard to be captured by the original growth method.

\begin{figure}[t]
    \centering
    \includegraphics[width=1\linewidth]{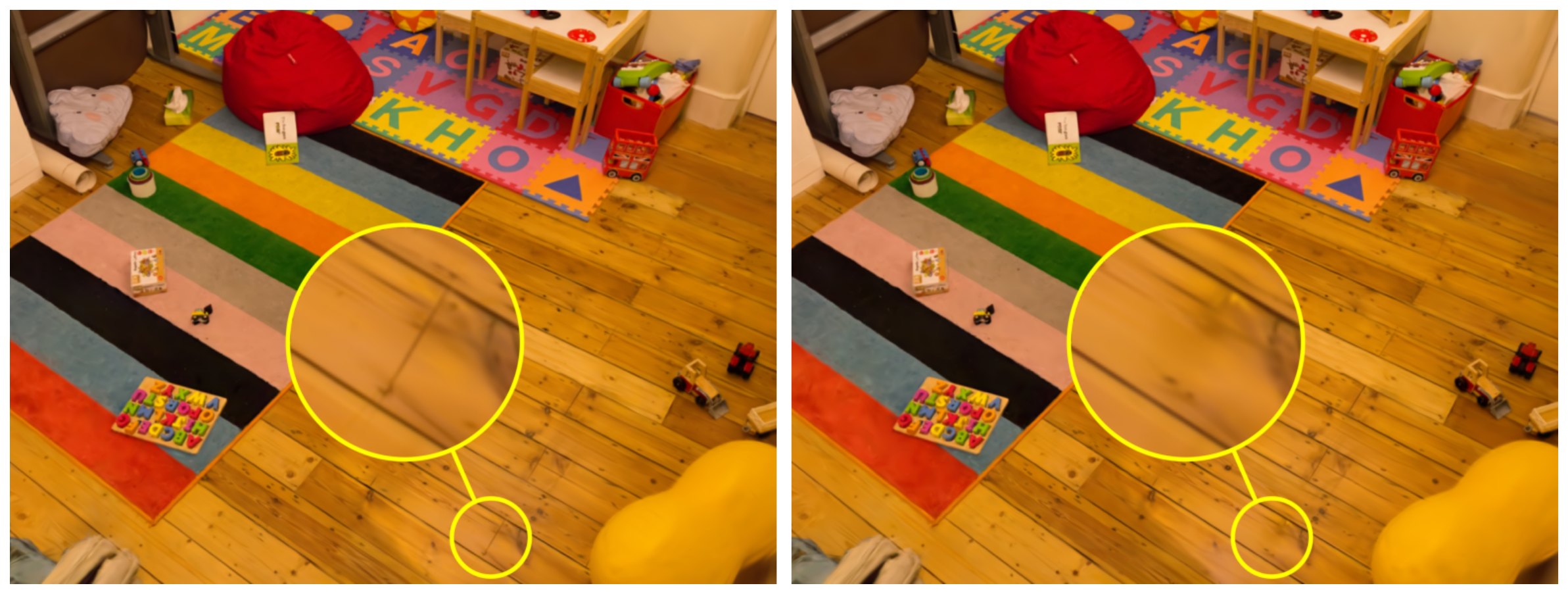}
    \vspace{-8mm}
    \caption{\textbf{Rendered images with our growing strategy.} (Left) Our proposed hybrid gradient-based method recovers the texture of the floor more accurately. (Right) Existing approaches based on position-only gradient lose details.  }
    \vspace{-2mm}
    \label{fig:rendered_image_diff_color_grad}
\end{figure}
\begin{figure*}[ht]
    \centering
    \includegraphics[width=0.99\linewidth]{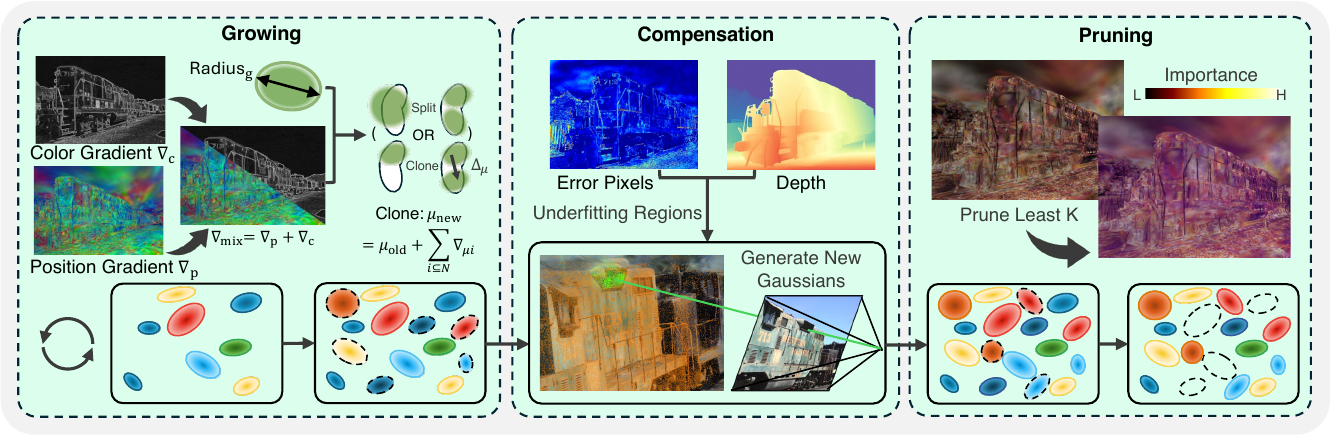}
    \vspace{-3mm}
    \caption{\textbf{Overall workflow of our proposed memory-bounded 3DGS training framework,} which iteratively performs growing, compensation, and pruning, progressively refining the representation capability.}
    \vspace{-4mm}
    \label{fig:overall_framework}
\end{figure*}
\noindent\textit{\Circled{3} \ul{Growing by position-only gradient cannot capture the blurry areas that need more Gaussians.}} 3DGS \cite{kerbl20233d} initializes the scene using a sparse point cloud generated from Structure-from-Motion (SfM) \cite{schonberger2016structure,ullman1979interpretation}, assigning default values to each Gaussian’s attributes. Then, it employs an adaptive density control algorithm to add new Gaussians during the densification step. In the densification, if the Gaussians' view-space positional gradients exceed a predefined threshold, they are candidates for duplication. Specifically, Gaussians with scales above a certain threshold are split, and otherwise are cloned. This strategy uses the view-space position gradient, computed via per-pixel color gradients, as an indicator for duplication. Referring to Equation \ref{eqn:render-color} and \ref{eqn:alpha}, we apply the chain rule to derive the gradient with respect to position $p_k$ of the $k$-th Gaussian, i.e.,
\begin{equation}\label{eqn:position-gradient}
\frac{d\ell}{dp_k}=\frac{d\ell}{dC}\frac{dC}{d\alpha_k}\frac{d\alpha_k}{dp_k},
\end{equation}
where $\ell$ is the rendering loss. Then, we analyze the second term, the partial derivative of $C$ with respect to $\alpha_k$, and expand it with other variables based on Equation \ref{eqn:render-color}. To be specific, this term can be represented by $\frac{dC}{d\alpha_k}=\sum_{j=k}^{N}-c_k\alpha_k\Pi(1-\alpha_k)$. This formulation shows that position gradients only capture partial color information and are influenced by the magnitude of the overlapped Gaussians' color. As a result, position gradients often fail to accurately detect underfit areas, particularly in blurred or low-contrast regions.

\textbf{Proposed solution:} According to Equation \ref{eqn:position-gradient}, we compute the gradient with respect to color $c_k$ of the $k$-th Gaussians, i.e., 
\begin{equation}\label{eqn:gradient-color}
\frac{d\ell}{dc_k}=\frac{d\ell}{dC}\frac{dC}{dc_k}.
\end{equation}
We also analyze the second term by expanding it as $\frac{dC}{dc_k}=\alpha_k\Pi_{j=1}^{k-1}(1-\alpha_j)$. It is seen that the gradient of color is not influenced by the other Gaussians' color and is only related to the transmittance. To address the aforementioned problem, we proposed to combine the two types of gradients, i.e., a mixture of both position and color gradients, to determine the areas where new Gaussians need to be added. Our investigation shows that our solution can accurately locate the blurry regions, allocating more primitives to complement the rendering quality, as shown in Fig. \ref{fig:rendered_image_diff_color_grad}.

\section{Proposed Training Framework}
\subsection{Framework Overview}

In the previous section, we have in-depth analyzed the limitations of existing 3DGS training methods, and we have introduced the motivations for our proposed framework. In this section, we will present our memory-efficient and high-quality training framework in detail. The overview of our proposed framework is illustrated in Fig. \ref{fig:overall_framework}. In summary, \ul{our framework dynamically grows, compensates, and prunes Gaussians in an iterative way, where those steps are alternately performed in every predefined iteration (e.g., every 50 iterations), progressively refining the representative capability under a consistent memory bound.} In the growing step, we clone and split Gaussians based on the proposed hybrid gradients criterion at a coarse-grained level. In other words, this step rapidly increases the number of Gaussians during the early densification, then our method continuously introduces new Gaussians at the rest of the clone step to refine the model by compensating for the representation capability in the low-quality areas. In the Gaussian compensation step, we fine-grainly identify the low-quality pixels with the highest error, project them back to their corresponding 3D location, and generate new Gaussians at that point to better capture underfitting regions. On the other hand, in the pruning phase, to ensure the model remains within a memory budget, we concurrently remove an equal number of less important Gaussians when the total count exceeds a predefined threshold. By iteratively performing these two steps, our framework adaptively discovers and optimizes a compact subset of Gaussians that preserves rendering fidelity while ensuring memory efficiency throughout training. The overall algorithm is presented in Algorithm \ref{alg:Overall_growth}. 

\subsection{Iterative Growing and Pruning}

The principle of our training framework is to alternately operate growing and pruning steps in every predefined iteration (e.g., every 50 iterations), dynamically adding informative Gaussians and removing redundant Gaussians. 
For the growing step, as discussed in section \ref{sec:motivation}, we incorporate the color gradient, $\nabla_\text{c}$, which reflects rendering error more accurately in the actual rendering space.
According to the gradient backpropagation formulation, Eq. \ref{eqn:position-gradient} and Eq. \ref{eqn:gradient-color}, color error flows entirely into the color gradients of individual Gaussians after being scaled by the transmittance weight. Therefore, we propose a mixed criterion that leverages color and position information to better identify regions requiring densification, i.e., $\nabla_\text{mix}=\nabla_\text{p}+\nabla_\text{c}$, to clone and split Gaussians.

As the comparison shown in Fig. \ref{fig:compensate_gaussians}, color gradients more reliably highlight underfitting regions, particularly in backgrounds and blurred areas. Our proposed hybrid metric provides a more informative cue for identifying and densifying under-structured geometric regions, as shown in Fig. \ref{fig:rendered_image_diff_color_grad}.

\textbf{Dynamical position adjustment. }
To address the overlapping problem in existing densification approaches that directly copy Gaussians, we propose an adaptive position adjustment method based on the accumulated gradient information to move new Gaussians to appropriate places. Specifically, we accumulate all position gradients over $N$ views, i.e.,
\begin{equation}\label{eqn:adjust-position}
\mu_{\text{new}} =\mu_{\text{old}} + \sum_{i\in N}\nabla_{\mu i}
\end{equation}
representing a stable and optimal position where new Gaussians should be. This light shift effectively resolves the overlapping problem that new Gaussians receive similar updating gradients.

As new Gaussians are added in the previous growing steps, an equal number of the least important Gaussians are subsequently removed in this pruning step, keeping the total peak training memory under the constraint. To achieve iterative pruning, it requires continuously identifying and removing the least important Gaussians. However, calculating a comprehensive importance score for each Gaussian can be computationally expensive. We apply the importance criterion \cite{niemeyer2024radsplat} with light calculation by aggregating the ray contribution of Gaussians $i$ along all rays of $N$ views. For Gaussians $G_i$, 
\begin{equation}\label{eqn:important-score}
R_i=\max_{ r \in R_f} \alpha_i^r \tau_i^r
\end{equation}
where $R_f$ represent all rays in the $N$ views, and $\tau_i=\alpha_i \prod_{j=1}^{i-1}(1-\alpha_j)$. This ray-based metric reflects the blending contribution of each Gaussian to the final pixel color and can be computed efficiently within the existing rendering pipeline, avoiding any significant overhead. Meanwhile, in our experiment, this process is performed every 100 iterations, resulting in negligible overhead.

In summary, our iterative growing and pruning have two advantages: firstly, it enables consistent training on devices with strict memory constraints where one-shot pruning approaches \cite{fan2023lightgaussian,fang2024mini} fail. Secondly, it allows the model to recover and re-optimize after each pruning, leading to a more balanced and high-quality sparse representation.

\begin{algorithm}[t!]
\newcommand{\assign}{$\leftarrow$}
\renewcommand{\algorithmicrequire}{\textbf{Input:}}
\newcommand{\smalltt}[1]{{\texttt{#1}}}
\caption{Overall procedure of our training framework.}\label{alg:Overall_growth}
\begin{algorithmic}[1]
\Require{Gaussian primitives $\bm{G}$, target peak number of Gaussians $F$, maximum number of iterations $T$;}

\If{\smalltt{densifyBegin} $<t<$ \smalltt{densifyEnd}}
    \If{$|\bm{G}|$ $<$ $F$}
        \State \smalltt{cloneAndSplit(}$\bm{G}$\smalltt{)}; 
        \State \smalltt{shiftNewGaussians($\bm{G}$)}; \hfill 
    \EndIf
    \State \smalltt{prune(}$\bm{G}, o_i<o_t$\smalltt{)};

\ElsIf{\smalltt{compensateBegin} $<t<$ \smalltt{compensateEnd}}
    \State \smalltt{compensateGaussians(}$\bm{G}$\smalltt{)}; \hfill 
\EndIf
\If{$|\bm{G}|$ $>$ $F$}
    \State $K=F-|\bm{G}|$; 
    \State $\bm{G}'$ \assign \smalltt{findLeastK(}$\bm{G},K$\smalltt{)}; 
    \State \smalltt{prune(}$\bm{G}, \bm{G}'$\smalltt{)};
\EndIf
\State $t=t+1$;
\end{algorithmic}
\end{algorithm}

\subsection{Adaptive Gaussian Compensation}

Positions of Gaussians receive only infinitesimal fluctuating gradients, updated by gradients propagated through the chain rule across varying camera views. As training progresses, the exponentially decaying learning rate further leaves the Gaussian stable, while reconstruction errors in underfit, blurry regions persist. To refine the poorly reconstructed and sparsely covered areas, we innovatively propose a Gaussian compensation approach before the pruning step to generate new Gaussians in the underfitting area based on a per-pixel error that measures the difference between the ground truth and the rendered image. Per-pixel error can be directly derived from the color gradient for each pixel, computed in the original backward pass, without incurring additional computational cost. 

Once underfitting pixels are identified, the next challenge is transforming 2D pixel coordinates into corresponding 3D Gaussian positions. We can replace the color $c_i$ of the $i$-th Gaussian with the depth of its center $d_i$ as 
\begin{equation}\label{eqn:render-depth}
  D=\sum_{i\in N}d_i\alpha_i \prod_{j=1}^{i-1}(1-\alpha_j).
\end{equation}
This approach approximately estimates the depth for each pixel. Then, inspired by \cite{fang2024mini}, we project the selected high-error pixels back into 3D space by replacing $d_i$ with $d_{\text{mid}}$, the Gaussian midpoints that contribute most to the pixel. For each image, we identify the top-$K$ pixels with the highest color gradient magnitude and generate new Gaussians at the corresponding positions derived from $\alpha$-blended depth. We set the colors of these Gaussians same as the ground truth pixel values. %

To be specific, as shown in Algorithm~\ref{alg:compensate-gaussian}, our Gaussian compensation identifies the top-$K$ pixels $\bm{p}_\text{err}$ with the highest errors and computes their corresponding 3D positions $\bm{D}_\text{err}$ in world space using depth alpha-blending. The ground truth color at each selected pixel, denoted as $\bm{p}_\text{gt}$, is also stored. After every user-defined interval, new Gaussians are generated at the computed positions $\bm{D}_\text{err}$, and each is assigned the corresponding color $\bm{p}_\text{gt}$. Fig. \ref{fig:compensate_gaussians} shows that the compensated Gaussians in our proposed approach recover the underfitting regions missed by the original 3DGS, improving reconstruction quality in blurry or sparse areas.

\begin{algorithm}[t!]
\newcommand{\assign}{$\leftarrow$}
      \renewcommand{\algorithmicrequire}{\textbf{Input:}}
      \newcommand{\smalltt}[1]{{\texttt{#1}}}
      \caption{The proposed Gaussian compensation.}
      \label{alg:compensate-gaussian}
      \begin{algorithmic}[1]
        \Require{Camera view $v$, ground truth image $\bm{I}_\text{gt}$}
        \State $\bm{I}$ \assign \smalltt{render(}$v$\smalltt{)}
        \State $\bm{p}_\text{err}$, $\bm{p}_\text{gt}$ \assign \smalltt{findErrorPixels(}$\bm{I},\bm{I}_\text{gt}$\smalltt{)}
        \State $\bm{D}_\text{err}$ \assign \smalltt{renderDepth(}$\bm{p}_\text{err}$\smalltt{)}
        \State \smalltt{list.append(}$\bm{D}_\text{err}, \bm{p}_\text{gt}$\smalltt{)}
        \If{$t$ \% \smalltt{compensateInterval} $=0$}
          \State \smalltt{genGaussians(list)}
          \State \smalltt{list.clear()}
        \EndIf
\end{algorithmic}
\end{algorithm}
\section{Evaluation}

\begin{table*}[ht!]
  \centering
\caption{\textbf{Quantitative results on multiple datasets, compared with existing state-of-the-art works.} Reducing-3DGS, Compact-3DGS and EAGLES results are replicated using official code. 3DGS, Mini-Splatting and Taming 3DGS results are reported from \cite{mallick2024taming}. ``\#G/M'' denotes the \textbf{peak} number of Gaussians in training (in millions). 
The \textbf{absolutely best results} are shown in bold, and the \toptwo{best results from efficient training methods} are highlighted. \greenbar{Horizontal bars} provide an intuitive comparison of the peak number of Gaussian points. ``$\downarrow$'' and ``$\uparrow$'' indicate lower and higher values are better, respectively.}
\vspace{-3mm}
    \resizebox{\linewidth}{!}{%
    \begin{tabular}{lcccc|cccc|cccc}
    \toprule
    \multirow{2}[2]{*}{Method} & \multicolumn{4}{c|}{Mip-NeRF 360} & \multicolumn{4}{c|}{Tanks\&Temples} & \multicolumn{4}{c}{Deep Blending} \\
\cmidrule{2-13}          & PSNR↑  & SSIM↑  & LPIPS↓ & \#G/M↓ & PSNR↑  & SSIM↑  & LPIPS↓ & \#G/M↓ & PSNR↑  & SSIM↑  & LPIPS↓ & \#G/M↓ \\
    \midrule
    3DGS \cite{kerbl20233d}  & 27.46 & 0.815 & 0.215 & 3.310 & 23.65 & 0.847 & 0.176 & 1.840 & 29.64 & 0.904 & 0.243 & 2.810 \\
    \midrule
    Mini-Splatting    \cite{fang2024mini} & 27.26 & \textbf{0.822} & \textbf{0.217} & \barChartA{4.320} & 23.42 & \textbf{0.847} & \textbf{0.181} & \barChartA{4.320} & \textbf{30.04} & \textbf{0.910} & \textbf{0.244} & \barChartA{4.510} \\
    Reducing-3DGS     \cite{papantonakis2024reducing}  & 27.21 & 0.811 & 0.225 & \barChartA{2.749} & 23.59 & 0.841 & 0.187 & \barChartA{1.507} & 29.61 & 0.903 & 0.248 & \barChartA{2.218} \\
    Compact-3DGS      \cite{lee2024compact}  & 26.96 & 0.797 & 0.244 & \barChartA{2.590} & 23.34 & 0.831 & 0.202 & \barChartA{1.465} & 29.80 & 0.900 & 0.257 & \barChartA{2.268} \\
    EAGLES            \cite{girish2024eagles}  & 27.15 & 0.811 & 0.231 & \barChartA{1.928} & 23.27 & 0.837 & 0.201 & \barChartA{0.954} & 29.83 & 0.909 & 0.246 & \barChartA{1.981} \\
    \midrule

    Taming 3DGS       \cite{mallick2024taming} &27.22 & 0.795 & 0.260 & \barChartA{0.632} & \toptwo{\textbf{23.68}}  & 0.836 & 0.211 & \barChartA{0.319} & 29.49 & 0.900 & 0.270 & \barChartA{0.294} \\
    \textbf{Ours} & \toptwo{\textbf{27.30}} & \toptwo{0.809} & \toptwo{0.234} & \barChartB{0.628} & 23.62 & \toptwo{0.842} & \toptwo{0.192} & \barChartB{0.318} & \toptwo{29.64} & \toptwo{0.906} & \toptwo{0.256} & \barChartB{0.292} \\

    \bottomrule
    \end{tabular}%
    }

    \vspace{-1mm}
  \label{tab:quantitative_comparision}%
\end{table*}%

\textbf{Dataset and metrics.} Following the standard practice, we evaluate our rendering performance on three novel view synthesis datasets: Mip-NeRF 360 \cite{barron2022mip}, Tank\&Temple \cite{Knapitsch2017}, and Deep-Blending \cite{hedman2018deep}. For quantitative evaluation, we report peak signal-to-noise ratio (PSNR), structural similarity (SSIM), and learned perceptual image patch similarity (LPIPS) \cite{zhang2018unreasonable}. Furthermore, we assess memory efficiency by measuring peak training memory usage on real-world edge settings, i.e., NVIDIA Jetson AGX Xavier. For visualized evaluation, we show rendering results of 3DGS \cite{kerbl20233d} and Taming 3DGS \cite{mallick2024taming} on various scenes for comparison.

\begin{figure*}[!ht]
    \centering
    \includegraphics[width=0.95\linewidth]
    {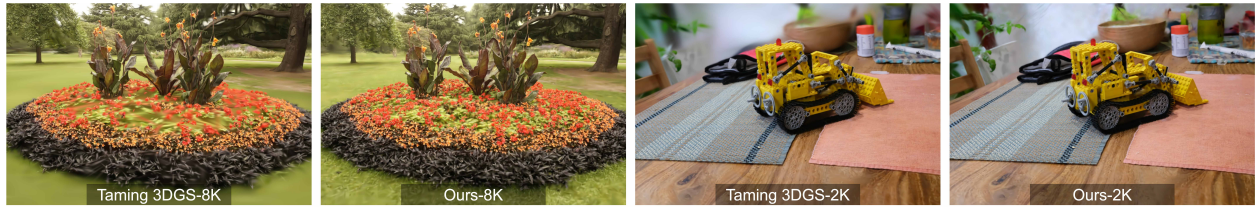}
    \vspace{-4mm}
    \caption{\textbf{Visualized results on flowers at the 8K-th iteration and kitchen at the 2K-th iteration.} Our method shows significantly improved rendering quality after the same training iterations compared to Taming 3DGS \cite{mallick2024taming}. }
    \vspace{-2mm}
    \label{fig:rendered_images_diff_iter}
\end{figure*}
 
\textbf{Implementation details.} All render quality experiments are conducted under the same environment specified in the original 3DGS \cite{kerbl20233d} and Taming 3DGS \cite{mallick2024taming} using an NVIDIA GTX 4090 GPU. Following Mini-Splatting \cite{fang2024mini}, we reset all Gaussians' opacity and position at the 5K-th iteration for Mip-NeRF 360 \cite{barron2022mip} outdoor scene. Our Gaussian compensation step starts at the 10K-th iteration and ends at the 15K-th iteration. After that, we fine-tune the result to a certain iteration depending on each scene.

\subsection{Quantitative Results}

Quantitative results are summarized in Table \ref{tab:quantitative_comparision}, in comparison with the original 3DGS \cite{kerbl20233d} and the state-of-the-art training method Taming 3DGS \cite{mallick2024taming}. We also compare to the state-of-the-art pruning works like Mini-Splatting \cite{fang2024mini}. It is seen that we outperform Taming 3DGS \cite{mallick2024taming} by an average of 0.15 dB PSNR and 0.03 LPIPS with fewer peak Gaussians across all scenes. Compared to the state-of-the-art pruning method, Mini-Splatting \cite{fang2024mini}, and the vanilla 3DGS \cite{kerbl20233d}, our method improves PSNR by 0.5 dB on the Tank\&Temple dataset and reduces peak numbers of Gaussians by more than 6$\times$.

\begin{figure}[t]
    \centering
    \includegraphics[width=1\linewidth]{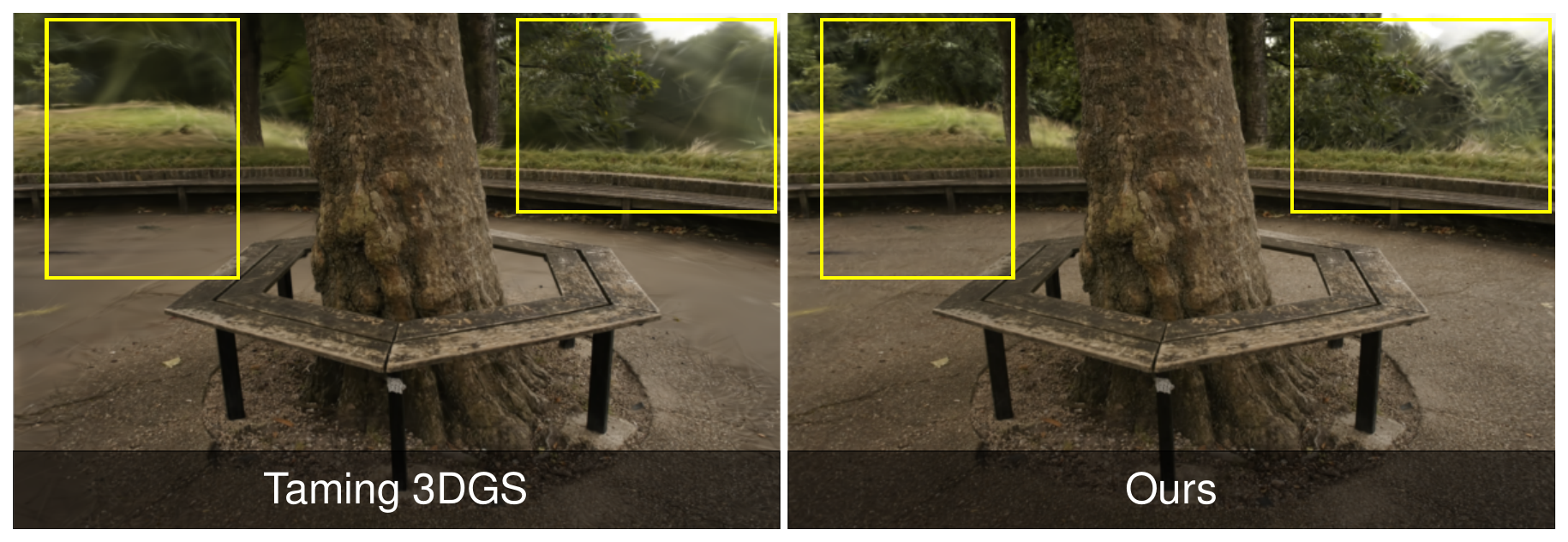}
    \vspace{-8mm}
    \caption{\textbf{Rendered image example.}  Our method presents significantly higher perceptual quality with high-frequency details, while Taming 3DGS \cite{mallick2024taming} shows blurry background trees and land.}
    \label{fig:rendered_image_treehill}
\end{figure}

More importantly, our method practically achieves on-device training on memory-constrained platforms. Experiments conducted on Jetson Xavier reveal our method reduces peak memory usage by nearly 2$\times$ compared to the original 3DGS, as shown in Table \ref{tab:ablation_memory}(b). Notably, we observed that up to three-quarters memory is used for dataset storage, we develop a parallel dataloader that dynamically prefetches and moves data between the storage and the memory according to the training pipeline. This effort further reduces peak training memory by more than 5 GB.

\subsection{Visual Quality Results}
We compare the rendered images for the perceptual analysis, as illustrated in Fig. \ref{fig:rendered_image_treehill}. It is seen that our method brings significantly higher visual quality with high-frequency and textural fidelity. On the contrary, Taming 3DGS \cite{mallick2024taming} loses details in textured regions such as tree bark and gravel surfaces. Moreover, Fig. \ref{fig:rendered_images_diff_iter} demonstrates that our actual rendering quality is superior to Taming 3DGS \cite{mallick2024taming} at the iterations. Those results further show the effectiveness of our iterative growing and pruning over the existing slow densification based on prediction.

\begin{figure*}[!ht]
    \centering
    \includegraphics[width=0.95\linewidth]
    {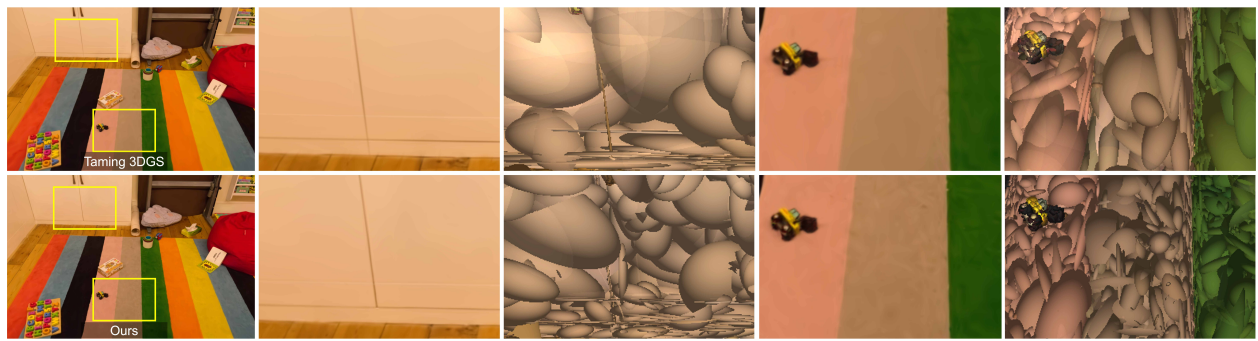}
    \vspace{-3mm}
    \caption{\textbf{Visualized ellipsoid results.} Our position adjustment reduces overlapped Gaussians (middle image), dynamically allocating more Gaussians to texture-rich regions (right image, texture of blanket), leading to a superior rendering quality.}
    \vspace{1mm}
    \label{fig:ellipsoid_compare}
\end{figure*}

We provide more results in \textbf{Appendix}. Fig. \ref{fig:rendered_image_all_p1} and Fig. \ref{fig:rendered_image_all_p2} show more rendered images on multiple scenes compared to Taming 3DGS \cite{mallick2024taming} and the original 3DGS \cite{kerbl20233d}. Our approach consistently achieves superior visual quality, particularly in high-frequency regions such as textured lawns and patterned blankets, where fine details are more faithfully preserved.
Additionally, we provide visualizations of Gaussian ellipsoids on the playroom scene to illustrate the spatial distribution of Gaussians, as shown in Fig.~\ref{fig:ellipsoid_compare}. Our method significantly reduces redundancy by eliminating excessive overlap among Gaussians (middle image) and dynamically allocates a higher density of Gaussians to texture-rich regions (right image), effectively capturing complex scene content.

\subsection{Ablation Study}

\begin{table}[t]
  \centering
  \caption{(a) LPIPS for the ``Playroom'' and ``Drjohnson'' scences. ``I.P.'' denotes ``Iterative Pruning'', ``G.C.'' denotes ``Gaussians Compensation''. (b) Peak training memory usage on NVIDIA Jetson Xavier. }
  \vspace{-4mm}
  \begin{minipage}{0.24\textwidth}
    \centering
    \caption*{(a) Ablation study}
    \vspace{-4mm}
    \resizebox{1\linewidth}{!}{
    \begin{tabular}{l|cc}
      \toprule
      Method & Playroom & Drjohnson \\
      \midrule
      Ours & 0.259 & 0.253 \\
      \cmidrule(lr){1-3}
      Baseline & 0.279 & 0.270 \\
      +I.P. & 0.264 & 0.257 \\
      +G.C. & 0.259 & 0.253 \\
      \bottomrule
    \end{tabular}
    }
    
  \end{minipage}
  \hspace{0.05\textwidth}
  \begin{minipage}{0.17\textwidth}
    \centering
    \caption*{(b) Mem usage (GB)}
    \vspace{-4mm}
    \resizebox{1\linewidth}{!}{
    \begin{tabular}{l|c}
      \toprule
      Method & Mem. \\
      \midrule
      3DGS & 18.59 \\
      Taming 3DGS & 10.01 \\
    \cmidrule(lr){1-2}  %
      Ours & \textbf{8.55} \\
      Ours w/loader & \textbf{2.98} \\
      \bottomrule
    \end{tabular}
    }
  \end{minipage}
  
  \label{tab:ablation_memory}
  \vspace{-1mm}
\end{table}

We conduct ablation studies to show the effectiveness of the components in our training framework. We test on the Deep-Blending \cite{hedman2018deep} dataset and report LPIPS scores to quantify the contribution of each component. We stop densification after Gaussians exceed a target number in the original 3DGS \cite{kerbl20233d} and report it as the baseline. Note that all configurations yield the same final number of Gaussians. 

Our first contribution involves an iterative pruning strategy, including hybrid gradients to grow new Gaussians and position adjustment after cloning the Gaussians. This procedure enables continual model refinement and yields an improvement of approximately 0.015 in LPIPS. Subsequently, we introduce the proposed Gaussian compensation, as illustrated in Fig.~\ref{fig:compensate_gaussians}, which results in a further LPIPS improvement of 0.005. This demonstrates that generating new Gaussians in the highest error pixel enhances perceptual fidelity in underfitting regions.

We also conducted an ablation study to evaluate the impact of using mixed gradients. The mixed gradient strategy yields a modest improvement on the test dataset (+0.04 dB PSNR) but shows a significant gain in training dataset (+0.87 dB PSNR) on the kitchen, indicating better realistic practical application. For additional ablation studies, please refer to our supplementary material.

\section{Conclusion}
We have presented a memory-efficient training framework for 3DGS that dynamically balances primitive growth and pruning under strict memory constraints. By iteratively refining Gaussians, coarse-grainly growing hybrid gradient varying areas, fine-grainly compensating underfitting regions while removing redundant ones, our approach achieves high-fidelity rendering with significantly reduced peak training memory consumption. As a result, our framework offers a scalable solution for deploying 3DGS under hardware constraints.
\FloatBarrier
{
    \small
    \bibliographystyle{ieeenat_fullname}
    \bibliography{main}
}

\clearpage
\setcounter{page}{1}
\maketitleappendix

\section{Additional Results}
\label{sec:Supplimentary}
\subsection{Additional Quantitative Results}
We summarize additional quantitative results on the Mip-NeRF 360, Tanks\&Temples, and Deep Blending datasets in Table \ref{tab:MipperScene}, Table \ref{tab:TanksTemplesAverAppendix}, and Table \ref{tab:DeepBlendingAverAppendix}.
\begin{table}[H]
    \centering
        \caption{Deep Blending per scene results. 3DGS results are reported from \cite{girish2024eagles}. Taming 3DGS \cite{mallick2024taming} results are replicated using official code.}
        \vspace{0mm}
    \resizebox{0.98\linewidth}{!}{%
        \begin{tabular}{cccccc}
        \toprule
        Scene & Method & PSNR↑  & SSIM↑  & LPIPS↓ & \#G/M$\downarrow$\\
        \midrule
        \multirow{4}{*}{Drjohnson} 
              & 3DGS  & 28.77 & 0.900 & 0.250 & 3.260 \\
              & Taming 3DGS  & \textbf{29.40} & 0.903 & 0.266 & 0.404 \\
              & Ours  & 29.33 & \textbf{0.904} & \textbf{0.253} & \textbf{0.400} \\
        \midrule
        \multirow{4}{*}{Playroom} 
              & 3DGS  & 30.07 & 0.900 & 0.250 & 2.290 \\
              & Taming 3DGS   & 29.59 & 0.898 & 0.274 & \textbf{0.185} \\
              & Ours  & \textbf{30.04} & \textbf{0.908} & \textbf{0.259} & \textbf{0.185} \\

        \midrule
        \multirow{4}{*}{Average} 
              & 3DGS  & 29.42 & 0.900 & 0.250 & 2.780 \\
              & Taming 3DGS   & 29.49 & 0.900 & 0.270 & 0.294 \\
              & Ours & \textbf{29.69} & \textbf{0.906} & \textbf{0.256} & \textbf{0.292} \\
        \bottomrule
        \end{tabular}%
    }

    \label{tab:DeepBlendingAverAppendix}
\end{table}
\begin{table}[H]
    \centering
        \caption{Tanks\&Temples per scene results. 3DGS results are reported from \cite{girish2024eagles}. Taming 3DGS \cite{mallick2024taming} results are replicated using official code.}
        \vspace{0mm}
    \resizebox{0.98\linewidth}{!}{%
        \begin{tabular}{cccccc}
        \toprule
        Scene & Method & PSNR↑  & SSIM↑  & LPIPS↓ & \#G/M$\downarrow$\\
        \midrule
        \multirow{3}{*}{Train} 
              & 3DGS  & 21.94 & 0.810 & 0.200 & 1.110 \\
              & Taming 3DGS   & 22.14 & 0.804 & 0.237 & \textbf{0.365} \\
              & Ours & \textbf{22.24} & \textbf{0.815} & \textbf{0.214} & \textbf{0.365} \\
        \midrule
        \multirow{3}{*}{Truck} 
              & 3DGS  & 25.31 & 0.880 & 0.150 & 2.540 \\
              & Taming 3DGS   & 25.22 & 0.868 & 0.184 & 0.272 \\
              & Ours & \textbf{25.00} & \textbf{0.869} & \textbf{0.170} & \textbf{0.270} \\
        \midrule
        \multirow{3}{*}{Average} 
              & 3DGS  & 23.63 & 0.850 & 0.180 & 1.830 \\
              & Taming 3DGS   & 23.68 & 0.836 & 0.211 & 0.319 \\
              & Ours & \textbf{23.62} & \textbf{.842} & \textbf{0.192} & \textbf{0.318} \\
        \bottomrule
        \end{tabular}%
    }

    \label{tab:TanksTemplesAverAppendix}
\end{table}
\begin{table}[H]
  \centering
    \caption{Mip-NeRF 360 per scene results. 3DGS results are reported from \cite{girish2024eagles}. Taming 3DGS \cite{mallick2024taming} results are replicated using official code.}
    \vspace{0mm}
     \resizebox{0.98\linewidth}{!}{%
  \begin{tabular}{lccccc}
    \toprule
    Scene & Method & PSNR $\uparrow$ & SSIM $\uparrow$ & LPIPS $\downarrow$ & \#G/M$\downarrow$ \\
    \midrule
    \multirow{3}{*}{Bicycle} & 3DGS & 25.13 & 0.750 & 0.240 & 5.310 \\
                      & Taming 3DGS & 24.85 & 0.718 & 0.295 & 0.813 \\
                             & Ours & \textbf{25.20} & \textbf{0.759} & \textbf{0.244} & \textbf{0.800} \\
    \midrule
    \multirow{3}{*}{Bonsai} & 3DGS & 32.19 & 0.950 & 0.180 & 1.250 \\
                     & Taming 3DGS & 31.86 & 0.936 & 0.220 & 0.413 \\
                            & Ours & \textbf{31.88} & \textbf{0.938} & \textbf{0.212} & \textbf{0.410} \\

    \midrule
    \multirow{3}{*}{Counter} & 3DGS & 29.11 & 0.910 & 0.180 & 1.170 \\
                      & Taming 3DGS & 28.59 & 0.898 & 0.223 & 0.311 \\
                            & Ours & \textbf{28.77} & \textbf{0.901} & \textbf{0.214} & \textbf{0.310} \\

    \midrule
    \multirow{3}{*}{Flowers} & 3DGS & 21.37 & 0.590 & 0.360 & 3.470 \\
                      & Taming 3DGS & 21.07 & 0.554 & 0.407 & 0.575 \\
                             & Ours & \textbf{21.16} & \textbf{0.590} & \textbf{0.354} & \textbf{0.570} \\

    \midrule
    \multirow{3}{*}{Garden} & 3DGS & 27.32 & 0.860 & 0.120 & 5.690 \\
                     & Taming 3DGS & \textbf{27.43} & 0.858 & 0.126 & \textbf{1.900} \\
                            & Ours & 27.36 & \textbf{0.865} & \textbf{0.107} & \textbf{1.900} \\

    \midrule
    \multirow{3}{*}{Kitchen} & 3DGS & 31.53 & 0.930 & 0.120 & 1.770 \\
                      & Taming 3DGS & 30.95 & 0.922 & 0.141 & 0.482 \\
                             & Ours & \textbf{31.35} & \textbf{0.924} & \textbf{0.135} & \textbf{0.480} \\

    \midrule
    \multirow{3}{*}{Room} & 3DGS & 31.59 & 0.920 & 0.200 & 1.500 \\
                   & Taming 3DGS & \textbf{31.27} & 0.908 & 0.250 & 0.225 \\
                          & Ours & 31.16 & \textbf{0.911} & \textbf{0.240} & \textbf{0.220} \\

    \midrule
    \multirow{3}{*}{Stump} & 3DGS & 26.73 & 0.770 & 0.240 & 4.420 \\
                    & Taming 3DGS & 26.01 & 0.735 & 0.293 & \textbf{0.480} \\
                           & Ours & \textbf{26.37} & \textbf{0.754} & \textbf{0.268} & \textbf{0.480} \\

    \midrule
    \multirow{3}{*}{Treehill} & 3DGS & 22.61 & 0.640 & 0.350 & 3.420 \\
                       & Taming 3DGS & \textbf{22.95} & 0.624 & 0.386 & 0.482 \\
                              & Ours & 22.48 & \textbf{0.635} & \textbf{0.329} & \textbf{0.480} \\

    \midrule
    \multirow{3}{*}{Average} & 3DGS & 27.45 & 0.810 & 0.220 & 3.110 \\
                      & Taming 3DGS & 27.22 & 0.795 & 0.260 & 0.632 \\
                             & Ours & \textbf{27.30} & \textbf{0.809} & \textbf{0.234} & \textbf{0.628} \\

    \bottomrule
  \end{tabular}
}
\label{tab:MipperScene}
\end{table}

\subsection{Additional Ablation Study}

Although our proposed framework primarily focuses on reducing training memory consumption, we also evaluate its impact on training speed. Specifically, we compare our method with the per-splat parallelized backpropagation used in Taming-GS. On the \textit{bicycle} scene, our approach reduces the training time from 10 minutes to 7 minutes, achieving a 30\% speedup. This improvement is attributed to more efficient memory usage and fewer redundant primitives during training, which reduces overhead in both forward and backward passes.

\subsection{Additional Visual Experiments}

\begin{figure*}[t]
    \centering
    \includegraphics[width=0.95\linewidth]
    {figs/rendered_images_all_p1_c.jpg}
    \vspace{-3mm}
    \caption{\textbf{Visualized results.} Our method achieves superior rendering quality compared against original 3DGS and Taming 3DGS \cite{mallick2024taming}.}
    \vspace{-2mm}
    \label{fig:rendered_image_all_p1}
\end{figure*}

\begin{figure*}[t]
    \centering
    \includegraphics[width=0.95\linewidth]
    {figs/rendered_images_all_p2_c.jpg}
    \vspace{-3mm}
    \caption{\textbf{Visualized results.} Our method achieves superior rendering quality compared against original 3DGS and Taming 3DGS \cite{mallick2024taming}.}
    \vspace{1mm}
    \label{fig:rendered_image_all_p2}
\end{figure*}

\end{document}